\documentclass[11pt]{article}

\usepackage[]{acl}

\usepackage{times}
\usepackage{latexsym}

\usepackage[T1]{fontenc}

\usepackage[utf8]{inputenc}

\usepackage{microtype}

\usepackage{inconsolata}

\usepackage{graphicx}

\usepackage{enumitem}
\usepackage{adjustbox}
\usepackage{amsmath}
\usepackage{amssymb}
\usepackage{amsthm}
\usepackage{cleveref}
\usepackage{stackengine}
\usepackage{comment}
\usepackage{mathtools}
\usepackage{xcolor}
\usepackage{longtable}
\usepackage{multirow}
\usepackage{graphicx}
\usepackage{subcaption}
\usepackage{colortbl}

\usepackage{booktabs}

\title{MoLAN: A Unified Modality-Aware Noise Dynamic Editing Framework for Multimodal Sentiment Analysis}

\author{
 \textbf{Xingle Xu},
 \textbf{Yongkang Liu},
 \textbf{Dexian Cai},
 \textbf{Shi Feng\thanks{Corresponding author.}},
\\
 \textbf{Xiaocui Yang},
 \textbf{Daling Wang},
 \textbf{Yifei Zhang}
\\
 Northeastern University, China,
\\
 \texttt{xuxingle@stumail.neu.edu.cn, misonsky@163.com, 2301840@stu.neu.edu.cn}
 \\ \texttt{\{fengshi, yangxiaocui, wangdaling, zhangyifei\}@cse.neu.edu.cn}
}

\begin{document}
\maketitle
\begin{abstract}
Multimodal Sentiment Analysis aims to integrate information from various modalities to make complementary predictions. However, it often struggles with irrelevant or misleading visual and auditory information. Most existing approaches treat entire modality as an independent unit for feature enhancement or denoising, which often suppresses redundant noise at the cost of weakening critical information. To address this challenge, we propose \textbf{MoLAN}, a unified \textbf{Mo}da\textbf{L}ity-aware noise dyn\textbf{A}mic editi\textbf{N}g framework. Specifically, MoLAN performs modality-aware block partitioning by dividing the features of each modality into multiple blocks. Each block is then dynamically assigned a distinct denoising strength based on its noise level and semantic relevance, enabling fine-grained noise suppression while preserving essential multimodal information. Notably, MoLAN is a unified and flexible framework that can be seamlessly integrated into a wide range of multimodal models. Building upon this framework, we further introduce \textbf{MoLAN\textsuperscript{+}}, a new multimodal sentiment analysis approach. Experiments across five models and four datasets demonstrate the broad effectiveness of the MoLAN framework. Extensive evaluations show that MoLAN\textsuperscript{+} achieves the state-of-the-art performance. The code is publicly available at \url{https://github.com/betterfly123/MoLAN-Framework}.

\end{abstract}

\section{Introduction}
Multimodal Sentiment Analysis (MSA) aims to integrate information from various modalities to achieve a more comprehensive and accurate understanding of the emotions \cite{zadeh2018multi,tsai2019multimodal}. MSA holds significant academic value in advancing multimodal learning, and offers broad industrial applications in areas such as human-computer interaction and mental health monitoring \cite{zhu2025integrating,singh2024deciphering}.

\begin{figure}[t]
  \centering
  \includegraphics[width=\linewidth]{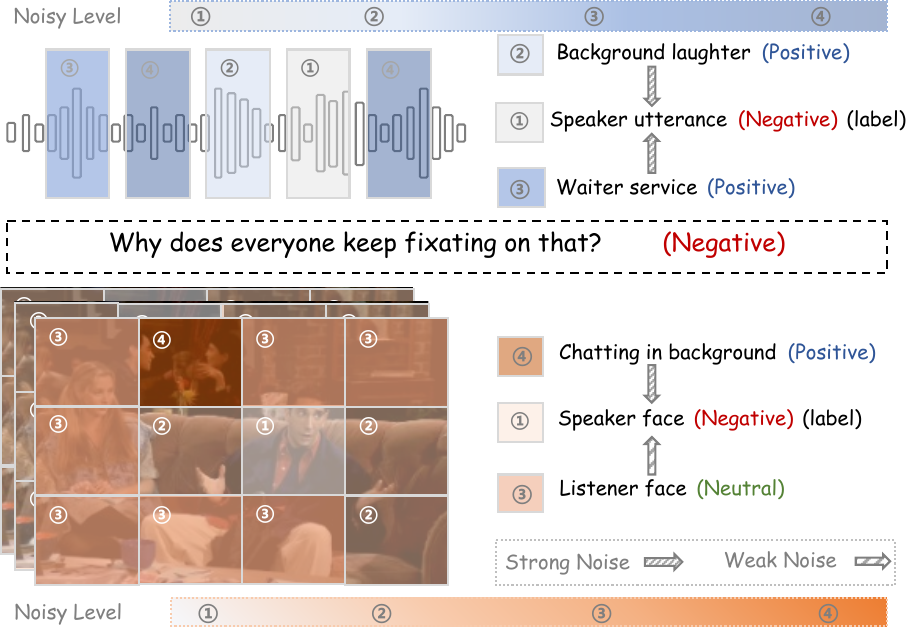}
  \caption{Distribution of noise. Lighter colors in the region mean more noise and less useful information.}
  \label{fig:intro}
\end{figure}

Existing MSA methods leverage multimodal synergy to achieve impressive improvements \cite{wu2024multimodal,li2025mpid}, but in real-world scenarios, noise interferes with representation learning and leads to performance degradation \cite{li2025t,liu2024noise}. To deal with multiple noise patterns, early solutions train individual models from scratch for each noise type \cite{yuan2021transformer} or design a unified model to perceive noise adaptively \cite{zeng2022robust}. However, noise sources differ substantially across modalities, making cross-modality noise transfer prone to failure. Accordingly, researchers design pattern specific denoising methods \cite{yuan2024meta}, yet they operate at the whole modality. Moreover, noise intensity varies across regions within the modality, so holistic denoising often suppresses noise at the cost of losing essential information.

As shown in Figure~\ref{fig:intro}, the intensity of this noise varies across different regions. Specifically, background smiles that contradict Ross’s emotional state constitute strong noise in the visual modality, while the remaining background mostly represents weak noise. In the audio modality, segments such as laughter that conflict with the sentiment label also form strong noise. The inconsistency of the noise distribution highlights the importance of fine-grained denoising. Therefore, the challenge of this paper is how to perform fine-grained noise dynamic editing on different modalities, so as to remove noise information while retaining information that is beneficial to MSA.

To address these issues, we propose \textbf{MoLAN}, a unified \textbf{Mo}da\textbf{L}ity-aware noise dyn\textbf{A}mic editi\textbf{N}g framework. To achieve dynamic fine-grained denoising, MoLAN employs a block partitioning strategy that divides each modality into different sub-blocks. In each block, the denoising strength is dynamically computed based on the noise level of the block. This approach allows the model to apply varying degrees of denoising across different blocks, thereby enhancing its ability of selective noise editing while preserving essential information in each modality. In addition, considering the heterogeneity between different modalities \cite{fan2024multi,wei2023tackling}, we design differentiated block partitioning strategies for each modality. Combining our experimental results and following the conclusions of previous studies \cite{li2025t,zhang-etal-2023-learning-language,lin2022multimodal}, we choose text modality as the main basis to calculate the denoising strength. Furthermore, MoLAN is a unified framework that can be flexibly integrated into different MSA models and Multimodal Large Language Models (MLLMs), thereby raising the upper limit of model performance. Based on the MoLAN, we propose \textbf{MoLAN\textsuperscript{+}}, which uses denoised information to update the cross-attention between modalities and guide the model to focus on important information for MSA. MoLAN\textsuperscript{+} further introduces denoising-driven contrastive learning to encourage the model to generate higher quality features, improving the performance of MSA task. The key contributions are as follows:
\begin{itemize}[leftmargin=*, nolistsep, noitemsep]
\item To address the noise, we propose \textbf{MoLAN}, a unified \textbf{Mo}da\textbf{L}ity-aware noise dyn\textbf{A}mic editi\textbf{N}g framework. It performs modality-aware block partitioning by dividing modality into multiple blocks. Each block is dynamically assigned a distinct denoising strength, enabling fine-grained noise editing. Additionally, MoLAN can be flexibly integrated into various models.
\item Based on the denoising framework, we further introduce the noise suppression cross-attention mechanism and denoising-driven contrastive learning, and design an MSA method \textbf{MoLAN\textsuperscript{+}}. MoLAN\textsuperscript{+} suppresses noise and guides the model to generate higher quality features.
\item We conduct experiments on seven models and four datasets to demonstrate the broad effectiveness of the MoLAN framework. Additionally, extensive evaluations on four benchmark multimodal datasets show that MoLAN\textsuperscript{+} achieves the state-of-the-art performance.
\end{itemize}

\section{Related Work}
\subsection{Multimodal Sentiment Analysis (MSA)}

MSA enables machines to understand emotions by leveraging visual, audio, and text signals. Early studies mainly adopt fusion methods such as TFN \cite{zadeh-etal-2017-tensor} and LMF \cite{liu2018efficient} to obtain joint representations. Subsequently, Transformer encoder architectures \cite{vaswani2017attention} and cross-modal attention become mainstream. For example, MulT \cite{tsai2019multimodal} uses cross-modal attention to align and fuse modalities, and related work \cite{zhou2025dual,wu2024multimodal,guo2024multimodal} further explores more effective alignment strategies. More recently, knowledge is also incorporated. KuDA \cite{feng2024knowledge} leverages affective knowledge to dynamically select the dominant modality and adjust modality contributions, while KEBR \cite{zhu2024kebr} injects non-verbal information from videos into textual semantics to enhance representations. Despite continuous progress in alignment and fusion, the impact of modality noise is often overlooked, which limits model performance. This work focuses on noise across modalities and performs noise editing to improve MSA.

\begin{figure*}[ht]
  \centering
  \includegraphics[width=\linewidth]{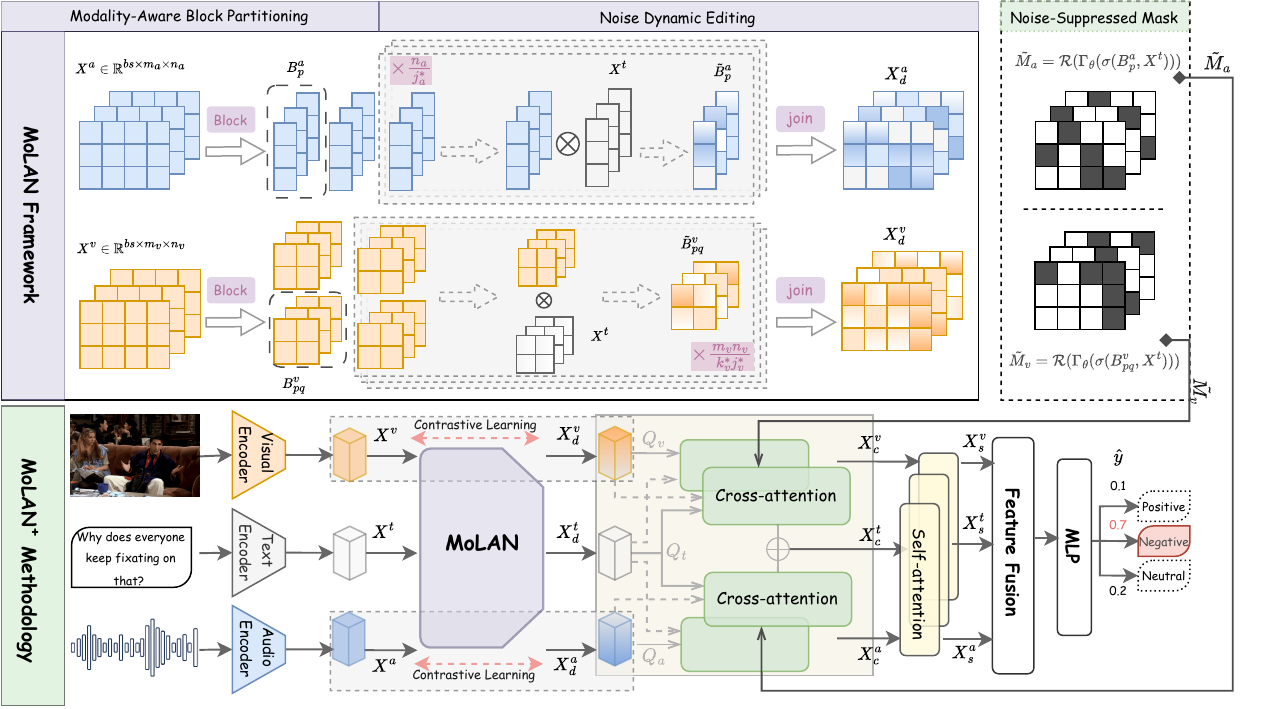}
  \caption{An illustration of MoLAN framework and MoLAN\textsuperscript{+} method. The purple box above represents the MoLAN framework, and the below represents the entire process of the MoLAN\textsuperscript{+} method. The MoLAN framework shown above provides a detailed description of the MoLAN block presented in  MoLAN\textsuperscript{+} method.}
  \label{fig:model}
\end{figure*}

\subsection{Multimodal Sentiment Analysis Denoising}

Recently, noise in MSA attracts increasing attention. t-HNE \cite{li2025t} removes visual and audio noise via text guidance and attention mechanisms. Meta-NA \cite{zeng2022robust} simulates noise tasks through meta-learning to improve robustness. JOSFD \cite{jiang2024joint} introduces fuzzy logic into multimodal fusion and decision-making to model emotion uncertainty. Missing modality is also regarded as a type of noise. EMMR \cite{zeng2022mitigating} reconstructs semantic features of key missing modalities, TATE \cite{zeng2022tag} uses tags to guide the model to focus on missing information, IASE \cite{shi2024information} aggregates data with a bipartite-graph formulation to reduce the impact of missing modalities, UMDF \cite{li2024unified} learns strong representations via a unified self-distillation mechanism, and Prompts are also used for missing modalities \cite{guo2024multimodal}. Unfortunately, existing works adopt too coarse processing granularity in the process of noise removal. This defect may cause excessive denoising and loss of essential information, or may lead to incomplete noise removal. In contrast, our work performs fine-grained noise dynamic editing on modality features, ensuring that essential information is preserved while denoising.


\section{MoLAN Framework}
As shown in the purple box in the upper part of Figure~\ref{fig:model}, the MoLAN framework consists of two components: modality-aware block partitioning and noise dynamic editing. The pilot study is provided in Appendix~\ref{sec:Pilot}.

\subsection{Modality-Aware Block Partitioning}

Since the distribution of noise is uneven, we introduce a block-level mechanism to enable fine-grained control over the denoising range. Through block partitioning, the minimum unit of the denoising operation shifts from the entire modality feature to a feature block, thereby achieving more precise denoising. Considering the differences between modalities, visual information usually presents in regional forms, which is suitable for two-dimensional block partitioning. In contrast, audio information appears as continuous segments, so one-dimensional block partitioning is more effective. Feature representation of modalities as \( X^f \in R^{bs \times m_f \times n_f}, f \in t, a, v\).
\begin{equation}
P_{block} =
\begin{cases}
\displaystyle (k^*_v, j^*_v), & \text{if } f = v \\
\displaystyle (j^*_a), & \text{if } f = a
\end{cases}
\end{equation}
where \(P_{block}\) is the optimal block partitioning parameter, it represents the size of the block. $bs$ is batch size. We use two-dimensional block partitioning as an example to illustrate the block partitioning process. \((k^*_v, j^*_v)\) is calculated as:
\begin{equation}
\begin{aligned}
  &(k^*_v, j^*_v) =  \\ &\underset{(k_v,j_v) \in \mathcal{D}_{m_v} \times \mathcal{D}_{n_v}}{\arg\min} \left\| \frac{k_v}{\sqrt{m_v}} - 1 \right\|_2^2 + \left\| \frac{j_v}{\sqrt{n_v}} - 1 \right\|_2^2 
\end{aligned}
\end{equation}
where \( \mathcal{D}_m \) and \( \mathcal{D}_n \) represent the sets of factors \( k_v \) and \( j_v \) of \( m_v \) and \( n_v \), respectively. The $\|\cdot\|_2^2$ represents the square of the L2 norm. Using the factor closest to the square root as the basis for block partitioning can achieve balanced segmentation: it can avoid information loss or noise residue caused by too large blocks, and semantic loss caused by too small blocks. The modality feature \( X^v \) is reshaped according to the block partitioning parameters, and each sub-block is defined as:
\begin{equation}
\begin{aligned}
B^v_{pq} &= X^v\left[(p-1)k^*_v,\; (q-1)j^*_v\right], \\
&\forall\; p \in \left[1, \frac{m_v}{k^*_v} \right],\;
q \in \left[1, \frac{n_v}{j^*_v} \right]\
\end{aligned}
\end{equation}
where \( B^v_{pq} \in R^{bs \times k^*_v \times j^*_v} \). \( p \) and \( q \) index the row and column positions of the sub-blocks. We conduct ablation studies to explain the choice of the block partitioning factor.

\subsection{Noise Dynamic Editing}
We first dynamically compute the adaptive denoising strength of each sub-block $B^v_{pq}$ and then edit its noise accordingly. The denoising strength is calculated as follows:
\begin{equation}
\sigma(B^v_{pq}, X^t) = \frac{\langle \Phi(B^v_{pq}), \Psi(X^t) \rangle}{\|\Phi(B^v_{pq})\|_2 \cdot \|\Psi(X^t)\|_2}
\end{equation}
where $X^t$ is text vector. $\Phi$ and $\Psi$ is a mapping function. $\sigma$ is the denoising strength for the block $B^v_{pq}$. Based on this strength, we perform dynamic editing. The denoising operation for each sub-block $B^v_{pq}$ can be represented as:
\begin{equation}
\tilde{B}^v_{pq} =  \sigma(B^v_{pq}, X^t) \cdot B^v_{pq}
\end{equation}
The denoised feature is obtained by recombining all blocks:
\begin{equation}
X_d^v = R(\{\tilde{B}^v_{pq}\}_{p=1,q=1}^{P,Q}) 
\end{equation}
where \( X_d^v \in R^{bs \times m_v \times n_v} \). $\mathcal{R}$ is the block reassembly operator. $P, Q$ is total block numbers. $P=m_v/k^*_v,Q=n_v/j^*_v$.

\section{MoLAN\textsuperscript{+} Methodology}

As shown in Figure~\ref{fig:model}, we propose the MoLAN\textsuperscript{+} method built upon MoLAN framework. The MoLAN\textsuperscript{+} method consists of three components: the MoLAN framework, noise-suppressed cross attention, and denoising-driven contrastive learning. The detailed introduction of each module can be found in following subsections.

\subsection{Problem Definition}

In MSA task, the input signal consists of $text (t)$, $visual (v)$ and $audio (a)$ modalities. The feature representation of these modalities can be denoted as \( X^f\in R^{bs \times m_f \times n_f}, f\in t, a, v \). The prediction is the sentiment score $\hat{y}$, which is a value.

To ensure a fair comparison, we use the same feature encoder as previous work \cite{tsai2019multimodal,wu2024multimodal, sun2025sequential}. After encoding, we feed the modality features into the MoLAN to obtain the denoised features:

\begin{equation}
X^v_d, X^a_d = MoLAN(X^t, X^v, X^a)
\end{equation}

\subsection{Noise-Suppressed Cross Attention}
In the green module below of Figure~\ref{fig:model}, to further enhance noise suppression, we update the attention mechanism based on the denoising strength calculation information. Take visual modality as an example:
\begin{equation}
M^v_{pq} = \Gamma_{\theta}(\sigma(B^v_{pq}, X^t)) \in \{0,1\}
\end{equation}
\begin{equation}
\Gamma_{\theta}(\sigma(B^v_{pq}, X^t)) = I(\sigma(B^v_{pq}, X^t) \geq \theta) \odot J_{pq}
\end{equation}
where $I(\cdot)$ is the indicator function. $J_{pq}$ is an all-ones matrix. $\theta \in [0,1]$ is the fixed similarity threshold. We aggregate the sub-blocks together to construct the mask matrix.
\begin{equation}
\tilde{M}_v = \mathcal{R}(\{M^v_{pq}\}_{p=1,q=1}^{P,Q}) 
\end{equation}
where $\mathcal{R}$ is the block reassembly operator that joins the sub-blocks according to their original positions. The mask matrix derived from the MoLAN is combined to calculate the cross-modality attention score to reduce the impact of noise.

\begin{equation}
X^{f_q}_c = \Omega_{\text{Inter-M}}(X^{f_q}_d, X^{f_{kv}}_d, \tilde{M}_{f_q})
\end{equation}
\begin{equation}
\begin{aligned}
\Omega_{\text{Inter-M}}&(X^{f_q}_d, X^{f_{kv}}_d, \tilde{M}_{f_q}) = \\ &\frac{exp_i[{X^{f_q}_d{X^{f_{kv}}_d}^T \cdot (\sqrt{d})^{-1} + \tilde{M}}_{mq}] X^{f_{kv}}_d}{\sum exp_j[{X^{f_q}_d{X^{f_{kv}}_d}^T \cdot (\sqrt{d})^{-1} + \tilde{M}_{mq}}]}
\end{aligned}
\end{equation}
where $f_q, f_{kv} \in \{t, a, v\}$. $f_q$ represents the query source modality of the current modality, and $f_{kv}$ represents the modality that provides key/value pairs. The text attention mask matrix is generated by the encoder. \(X^{f_q}_c\) represents the feature after the attention mechanism.

\subsection{Denoising-Driven Contrastive Learning}
As shown in the light gray below of Figure~\ref{fig:model}, to enhance the encoder's capability in distinguishing noise, we introduce a noise-driven contrastive learning loss. We perform contrastive learning between the denoised modal features and their corresponding original features. By minimizing the distance between positive pairs and maximizing the differences with other samples, the model is encouraged to learn more discriminative denoised representations. The contrastive learning loss function can be formulated as follows:

\begin{equation}
\begin{aligned}
& \mathcal{L}_{\text{contrast}} =  \\
& - E \log 
\left[
\frac{
\exp\left(\phi(X^v_d, X^v) / \tau\right)
}{
\sum_{j=1}^{N} \exp\left(\phi(X^v_d, X^v_j) / \tau\right)
}
\right] \\
& - E \log 
\left[
\frac{
\exp\left(\psi(X^a_d, X^a) / \tau\right)
}{
\sum_{j=1}^{N} \exp\left(\psi(X^a_d, X^a_j) / \tau\right)
}
\right]
\end{aligned}
\end{equation}
where $\phi$ and $\psi$ denote similarity measurement functions, with cosine similarity adopted in this work. $\tau$ is a temperature parameter that controls the smoothness of the distribution. During training, this module guides the encoder to focus on distinguishing modal noise, thereby improving the overall denoising quality and the effectiveness of fusion.

\subsection{Sentiment Prediction}
By introducing a cross-attention mechanism, information from different modalities is effectively interacted and integrated, enabling the initial fusion of multimodal features. Subsequently, the interacted representations are fed into a self-attention mechanism to further model the deep intra-modal dependencies. Finally, the information from different modalities is consolidated into a unified representation, which is used for the sentiment analysis.
\begin{equation}
X^f_s = SelfAttn(X^{f}_c, \tilde{M}_{f})
\end{equation}
\begin{equation}
\hat{y} = \text{MLP}_{\theta_{\text{FC}}} \left( \mathcal{F}_{\text{fuse}} \left( \text{Cat} \left( X^{t}_s, X^{a}_s, X^{v}_s \right) \right) \right)
\end{equation}
where $\theta_{\text{FC}}$ denotes the parameters of the fully connected network. $\mathcal{F}_{\text{fuse}}$ denotes linear layer, and $\hat{y}$ is the predicted sentiment value. The overall training of the MoLAN\textsuperscript{+} is performed by minimizing the following loss:
\begin{equation}
\mathcal{L} = \mathcal{L}_{\text{task}} + \mathcal{L}^v_{\text{contrast}} + \mathcal{L}^a_{\text{contrast}}
\end{equation}
where $\mathcal{L}_{\text{task}}$ involves regression and classification tasks. For regression tasks, we adopt the L1 loss, following prior works \cite{mai2023learning, mai2020modality}, which measures the absolute difference. For classification tasks, the standard cross-entropy loss is employed to optimize the model. The loss of the predicted value $\hat{y}$ and the ground truth $y$ is:
\begin{equation}
\mathcal{L}_{\text{task}} = 
\begin{cases}
\mathcal{L}_{\text{reg}} = \frac{1}{N} \sum\limits_{i=1}^{N} | y_i - \hat{y}_i | & \text{(Reg.)} \\
\mathcal{L}_{\text{cla}} = \frac{1}{N} \sum\limits_{i=1}^{N} -y_i \log(\hat{y}_i) & \text{(Cls.)}
\end{cases}
\end{equation}
where $N$ is the number of the samples. The model is trained based on the overall loss function.

\section{Experiments}

\begin{table*}[ht]
    
    \begin{adjustbox}{max width=\textwidth, center}
        \includegraphics[width=\textwidth]{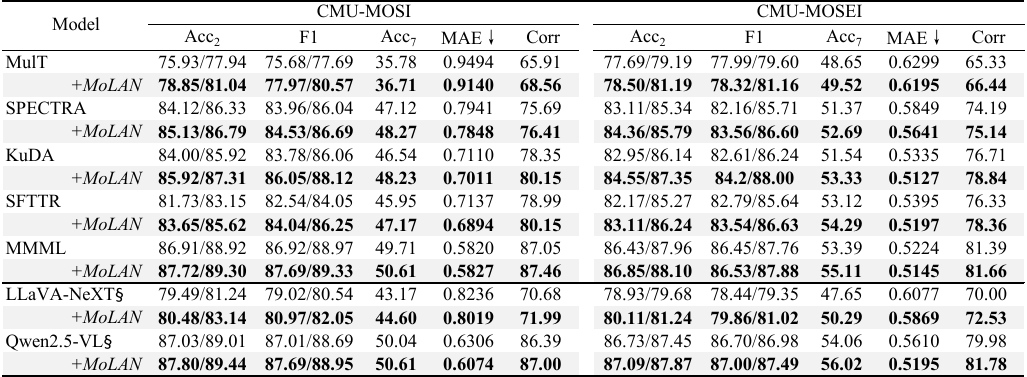}
    \end{adjustbox}
    \caption{The performance of the MoLAN framework on the MOSI and MOSEI. The baseline results in the experiment are obtained through replication. Two evaluation metrics, ACC and F1, are adopted, specifically ACC$_{2Has0}$ / ACC$_{2Non0}$ and F1$_{Has0}$ / F1$_{Non0}$. $\S$ denotes fine-tuning with LoRA. We perform significance testing on seven experimental groups, with p-value of \( 5.33 \times 10^{-5} \), \( 1.43 \times 10^{-6} \), \( 4.87 \times 10^{-10} \), \( 2.68 \times 10^{-6} \), \( 2.16 \times 10^{-3} \), \( 2.45 \times 10^{-5} \), and \( 2.66 \times 10^{-6} \), all of which \(<\) 0.05 indicate significant differences.}
    \label{tab:framework_1}
\end{table*}
\begin{table}[h]

  \includegraphics[width=\columnwidth]{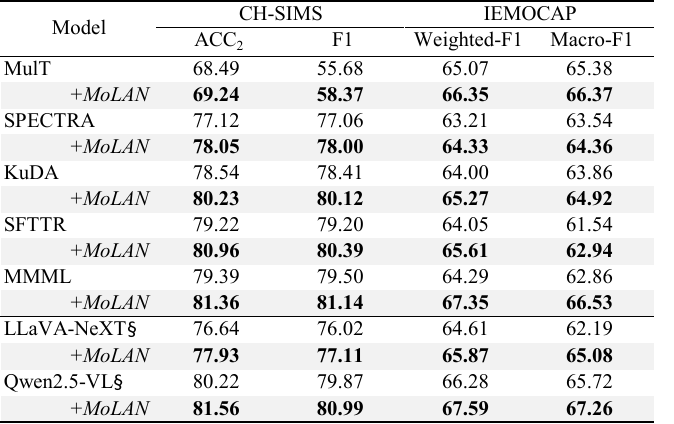}
\caption{The performance of the MoLAN framework on the SIMS and IEMOCAP. $\S$ denotes fine-tuning with LoRA. Significance testing in Appendix~\ref{app:st}.
}
\label{tab:framework_2}
\end{table}
\subsection{Datasets and Baselines}

We use CMU-MOSI \cite{Zadeh2016MOSIMC}, CMU-MOSEI \cite{Zadeh2018MultimodalLA}, CH-SIMS \cite{Yu2020CHSIMSAC}, and IEMOCAP \cite{Busso2008IEMOCAPIE}. 

\noindent\textbf{Framework:} We compare our method with \textbf{MulT} \cite{tsai2019multimodal}, \textbf{SPECTRA} \cite{yu-etal-2023-speech}, \textbf{KuDA} \cite{feng2024knowledge}, \textbf{SFTTR} \cite{sun-tian-2025-sequential}, \textbf{MMML} \cite{wu2024multimodal}, as well as MLLM-based baselines \textbf{LLaVA-NeXT} \cite{li2024llava} and \textbf{Qwen2.5-VL} \cite{bai2025qwen2}.

\noindent\textbf{Models:} We further benchmark against representative MSA methods, including \textbf{TFN} \cite{zadeh-etal-2017-tensor}, \textbf{LMF} \cite{liu-etal-2018-efficient-low}, \textbf{MFM} \cite{tsai2018learning}, \textbf{Self-MM} \cite{Yu_Xu_Yuan_Wu_2021}, \textbf{UniMSE} \cite{hu-etal-2022-unimse}, \textbf{CHFN} \cite{10.1145/3503161.3548137}, \textbf{ALMT} \cite{zhang-etal-2023-learning-language}, \textbf{EMT} \cite{sun2023efficient}, \textbf{GLoMo} \cite{10.1145/3664647.3681527}, \textbf{JOSFD} \cite{jiang2024joint}, \textbf{t-HNE} \cite{li2025t}, and \textbf{MMML} \cite{wu2024multimodal}. More details are in Appendix~\ref{app:details}.

\subsection{Framework Overall Analysis}

This section provides an overall analysis of the proposed framework to assess its performance across different evaluation dimensions. We first examine the framework’s effectiveness in enhancing MSA, followed by an analysis of its universality ability.

\begin{table*}[t]

    \begin{adjustbox}{max width=\textwidth, center}
        \includegraphics[width=\textwidth]{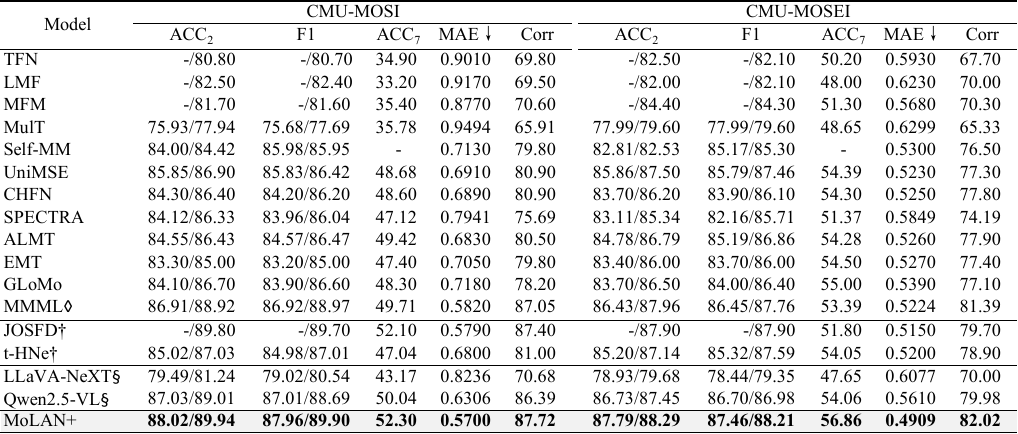}
    \end{adjustbox}
      \caption{The performance of the MoLAN\textsuperscript{+} on the MOSI and MOSEI. $\lozenge$ indicates our reproduced results. $\dagger$ indicates MSA denoising method. $\S$ denotes fine-tuning with LoRA. We perform significance testing on MMML and MoLAN\textsuperscript{+}, with p-value of \( 7.52 \times 10^{-4} \) \(<\) 0.05 indicates significant differences. The best results are in bold.}
  \label{tab:method_1}
\end{table*}

\begin{table}[]

\includegraphics[width=\columnwidth]{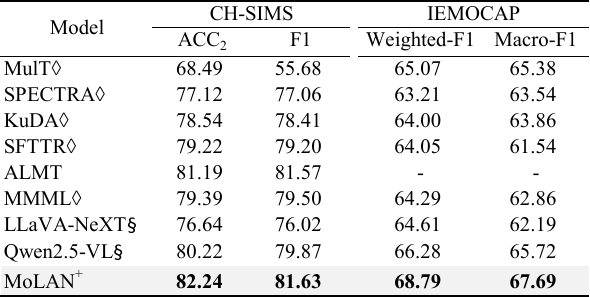}
\caption{Performance of MoLAN\textsuperscript{+} on the SIMS and IEMOCAP.  Significance testing in Appendix~\ref{app:st}.}
\label{tab:method_2}
\end{table}

\noindent\textbf{Framework effectiveness.} We validate the proposed MoLAN framework on four standard MSA datasets, using five representative MSA models and two MLLMs. Detailed experimental results are shown in Table~\ref{tab:framework_1} and Table~\ref{tab:framework_2}. Five MSA models all show dramatically performance improvements after integrating the MoLAN framework. This trend demonstrates that MoLAN can be universally applied across diverse model architectures, effectively improving the quality of multimodal feature fusion. This further demonstrates that MoLAN, through fine-grained noise editing of modality information, can suppress irrelevant noise while retaining key information, thereby strengthening semantic alignment and collaborative representation across modalities. It is worth noting that even the current SOTA, MMML, achieves further improvement after integrating MoLAN. This observation indicates that existing methods still have limitations in modality denoising, and that MoLAN helps overcome these performance bottlenecks. Furthermore, experiments on MLLMs further validate the framework's effectiveness. Both LLaVA-NeXT and Qwen2.5-VL exhibit consistent performance improvements after integrating MoLAN. Result demonstrates that MoLAN is not only applicable to traditional MSA models but also seamlessly integrates with mainstream MLLM architectures.

\noindent\textbf{Framework universality.} Experimental results show that whether MoLAN is integrated into dedicated MSA models or deployed on MLLMs, it consistently exhibits significant performance improvements. This consistent trend verifies the strong universality capability of MoLAN, indicating that it effectively adapts to different architectures, task settings, and data distributions. Overall, the results suggest that MoLAN serves as a general enhancement module for MSA tasks, providing a reliable solution for multimodal representation learning.

\begin{figure}[t]
  \centering
  \includegraphics[width=\linewidth]{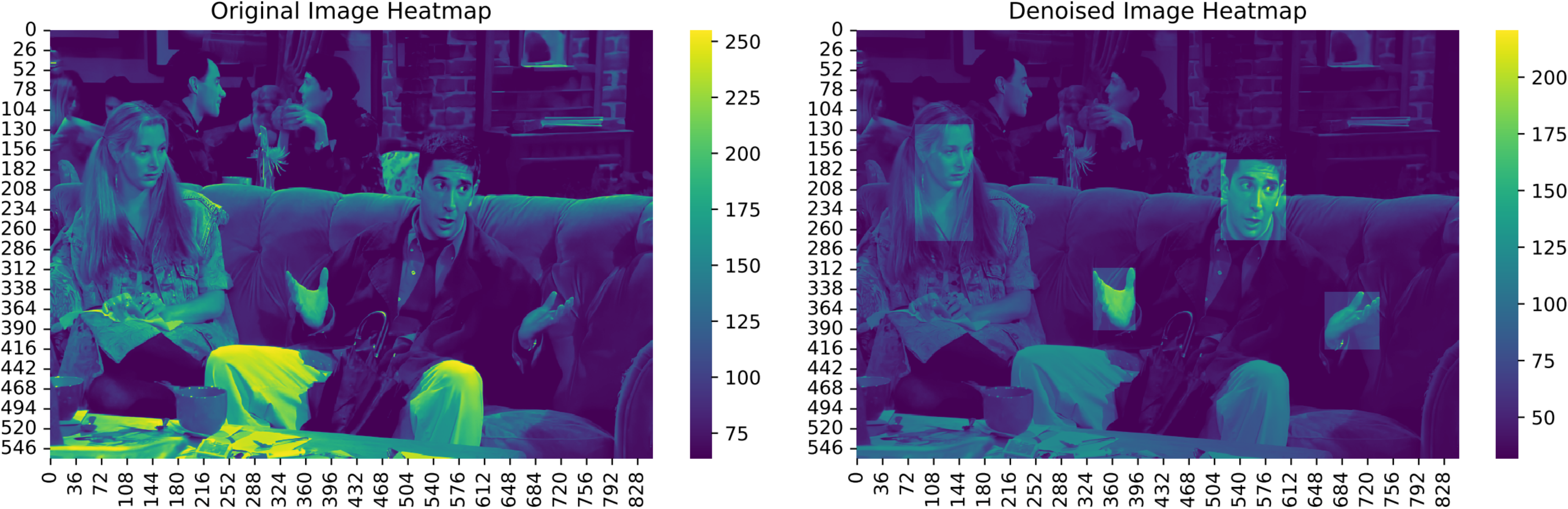}
  \caption{Pixel-level heatmap. Color intensity indicates the magnitude of the pixel value, with brighter areas representing stronger image information.}
  \label{fig:p_heatmap}
\end{figure}

\subsection{Effectiveness of Dynamic Strategy}
The superior performance of the framework demonstrates the effectiveness of our proposed noise dynamic editing. Next, we illustrate the key role of noise dynamic editing from the perspectives of visualization. Figure~\ref{fig:p_heatmap} shows the heatmap comparison between the original image and the image processed by the MoLAN framework. It can be clearly observed that the energy distribution of the heatmap of the original image is uniform, and the emotional information (the target person's face and hands) is not prominent enough due to the noise in the background area. After the noise dynamic editing of the MoLAN framework, the energy distribution of the heatmap on the right changes significantly. For the background noise area, the overall energy level is reduced, effectively suppressing the noise. In contrast, energy levels in areas such as the face remain high, forming clear highlights. This phenomenon intuitively illustrates that the MoLAN framework can dynamically control the area and intensity of noise editing, effectively removing noise while retaining essential information.

\subsection{MoLAN\textsuperscript{+} Experiments}
We conduct a comprehensive evaluation of MoLAN\textsuperscript{+} on four datasets. As presented in Table~\ref{tab:method_1} and Table~\ref{tab:method_2}, MoLAN\textsuperscript{+} consistently achieves SOTA performance across all datasets, surpassing both existing baseline models and recent MLLM-based approaches. This superior performance demonstrates that the noise-suppression cross attention and noise-driven contrastive learning modules effectively enhance the model’s discriminative capability and semantic alignment across modalities. By emphasizing emotion-relevant multimodal representations during feature extraction, the model is able to mitigate the influence of noisy information and maintain stable performance across diverse conditions. Moreover, when compared with the framework experiments, we observe that although other models benefit from the integration of the MoLAN framework, their results still fall short of the overall performance achieved by MoLAN\textsuperscript{+}. These findings suggest a strong synergy between noise-suppression cross attention, noise-driven contrastive learning, and the MoLAN framework, jointly contributing to the superior adaptability and effectiveness of MoLAN\textsuperscript{+} in MSA tasks.

We conduct a comparison between MoLAN\textsuperscript{+} and two representative denoising-based MSA models, JOSFD and t-HNE.
The significance tests yield p-values of 0.01 and 4.85$\times10^{-5}$, both of which are far below the 0.05 threshold, indicating that the performance improvement achieved by MoLAN\textsuperscript{+} is statistically significant. This result suggests that traditional denoising strategies often over-filter multimodality signals, inadvertently removing critical semantic information and thereby degrading model performance. In contrast, MoLAN\textsuperscript{+} employs a noise dynamic editing mechanism that selectively suppresses irrelevant noise while preserving emotion-relevant features, leading to higher accuracy across multiple datasets.

\subsection{Ablation Study}

\begin{table}[t]

\includegraphics[width=\columnwidth]{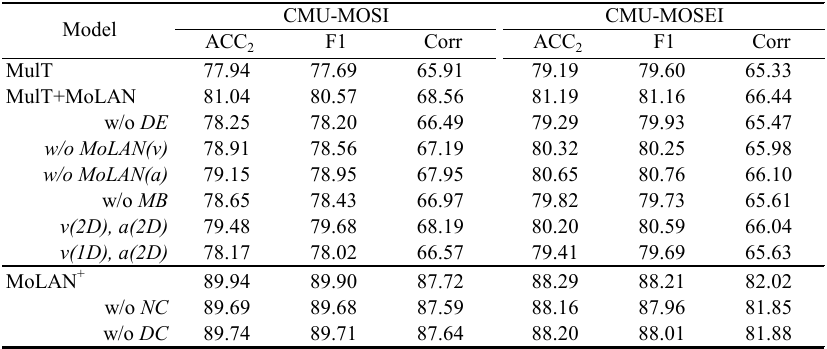}
\caption{Ablation studies.}
\label{tab:ablation_studies}
\end{table}
As shown in the upper part of Table~\ref{tab:ablation_studies}, we first conduct ablation experiments on two parts of MoLAN.

\noindent \textbf{Effectiveness of noise dynamic editing.} We apply a uniform denoising strength to the modality information. Experimental results (w/o DE) show that the model performance degrades after removing dynamic editing. This indicates that uniform denoising strength cannot handle the differences in noise levels between different regions. Such denoising may result in insufficient denoising in high-noise areas, while low-noise areas may lose valuable information due to over-denoising. The experimental results further verify the necessity of dynamic editing, which adaptively adjusts denoising strength based on local noise characteristics to achieve a better balance between noise suppression and information preservation.

We independently apply the denoising framework to the visual modality (MoLAN(v)) and the audio modality (MoLAN(a)). The results show that although the single-modality denoising does not reach the performance level of joint multimodal denoising, it still achieves a significant improvement compared with the model without the denoising framework. This finding indicates that the denoising mechanism also plays a positive role in single-modality denoising, while the collaborative removal during joint multimodal denoising further enhances the overall performance.

\noindent \textbf{Effectiveness of modality-aware block partitioning.} We conduct an ablation study on the modality-aware block partitioning strategy to examine its effectiveness. Specifically, we first apply a uniform one-dimensional block partitioning to all modalities (w/o MB), and then a uniform two-dimensional block partitioning to both the visual and audio modalities (v(2D), a(2D)). The results show a noticeable decline in performance, indicating that the information distribution differs significantly across modalities. These findings suggest that only a targeted block partitioning strategy can effectively localize and suppress modality-specific noise. Furthermore, when we assign one-dimensional block partitioning to the visual modality and two-dimensional block partitioning to the audio modality (v(1D), a(2D)), the model performance also decreases. This further verifies that the adopted configuration, namely 2D block partitioning for visual and 1D block partitioning for audio, better aligns with the properties of each modality and thus enables more efficient denoising.

As shown in the lower part of Table~\ref{tab:ablation_studies}, we perform ablation experiments for MoLAN\textsuperscript{+}. 

\noindent \textbf{Effectiveness of noise-suppressed cross attention.} We remove the noise-suppression cross-attention mechanism (w/o NC) to examine its effect. The experimental results show a performance drop, indicating that denoising information plays a crucial role in cross-attention computation. Specifically, when the mask matrix is removed, the model’s ability to perceive and suppress noise weakens, leading to an overall decline in performance. These results suggest that explicitly incorporating denoising information into the mask matrix effectively guides the cross-attention mechanism to focus on informative regions, reduce noise interference, and enhance the model’s noise resistance.

\noindent \textbf{Effectiveness of denoising-driven contrastive learning.} We further remove the denoising-driven contrastive learning module (w/o DC) to validate its contribution. The results reveal a drop in performance, suggesting that this module is essential for learning stable and discriminative modality representations. In particular, the contrastive objective establishes a semantic constraint between denoised and original features, enabling the model to maintain both distinctiveness and consistency of feature distributions under noisy conditions. Once this mechanism is removed, such relational constraints are weakened, making it difficult for the model to separate noise from informative signals in the latent space. Therefore, the denoising-driven contrastive learning not only improves cross-modal alignment but also enhances the model’s universality ability in complex noisy environments. Additional analyses and case studies in Appendix~\ref{app:more} and Appendix~\ref{app:case}.

\section{Conclusion}


We propose MoLAN, a unified modality-aware noise dynamic editing framework that partitions modalities into blocks and dynamically assigns denoising strength based on each block’s noise level and semantic relevance. MoLAN is plug-and-play and can be integrated into various MSA models and MLLMs to improve performance. Built on MoLAN, MoLAN\textsuperscript{+} further introduces noise suppression cross-attention mechanism and denoising-driven contrastive learning to emphasize essential information. Extensive experiments demonstrate the effectiveness of MoLAN and MoLAN\textsuperscript{+}. Overall, our work provides a practical pathway for enhancing robustness in multimodal systems.


\section*{Limitations}

Our current evaluation mainly covers several representative MSA benchmarks to demonstrate the generality and integrability of MoLAN and MoLAN\textsuperscript{+}. However, we do not yet conduct systematic evaluations under more challenging settings, such as cross-domain generative tasks or more complex real-world scenarios. In these settings, noise patterns and cross-modal interaction modes can be more diverse, thereby posing new requirements for noise editing. Therefore, comprehensive experiments and analyses on these broader scenarios remain to be further complemented.

\bibliography{custom}
\newpage
\appendix

\section{Pilot Study}
\label{sec:Pilot}

\begin{table}[t]

\includegraphics[width=\columnwidth]{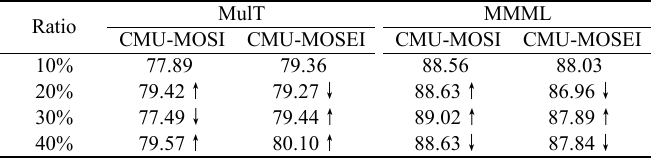}
\caption{Multimodal Noise. ${\downarrow}$ indicates a performance drop compared to the previous row, while ${\uparrow}$ indicates an improvement. The metric used is ACC$_2$.}
\label{tab:mul_noise}
\end{table}

The initial modality feature embedding, serving as the encoded representations of each modality, form the foundation of MSA. However, existing studies generally overlook the denoising of these encoded features and fail to investigate the potential impact of noise within the initial feature representations. Noise within the initial features may interfere with the accuracy of MSA and hinder the model’s ability to capture critical information. By enhancing the quality of the initial features through effective denoising, the overall performance of the model can be further improved. Therefore, we design and conduct a series of validation experiments to systematically analyze the presence and influence of noise in the initial modality features.

\subsection{Random Masking Strategy}

To investigate the impact of noise in the initial multimodal feature representations, we design a random masking strategy. Specifically, we randomly mask a portion of the elements in the feature embeddings according to a predefined masking ratio. In this way, we observe how the model performs under conditions of partial information loss.
\begin{equation}
M_{ij} = 
\begin{cases}
0, & \text{if } R_{ij} < p \\
1, & \text{otherwise}
\end{cases}
\quad R_{ij} \sim \mathcal{U}(0, 1)
\end{equation}
\begin{equation}
\tilde{\mathbf{F}} = \mathbf{F} \odot \mathbf{M}
\end{equation}
where \( \mathbf{F} \) denote the feature embedding matrix, and \( \mathbf{M} \) the corresponding mask matrix. Each element \( R_{ij} \) is a random number sampled from a uniform distribution over the interval \([0, 1]\). The masking ratio is defined by \( p \in [0, 1] \).

If the initial feature information contains little noise, an intuitive inference is that the model performance will degrade as the masking ratio increases. Therefore, we gradually increase the masking ratio and observe the trend in model performance, thereby indirectly reflecting the level of potential noise in the initial multimodal features.

\subsection{Multimodal Noise}

We first apply the random masking strategy to the overall modality-level features and conduct experiments on both the visual and audio modalities with the same masking ratio. As shown in Table~\ref{tab:mul_noise}, the results show that the model performance improves as the masking ratio increases. This phenomenon indicates that the removed portions of the features may contain more noise than useful information. Therefore, eliminating these noisy components allows the model to focus on more critical representations, leading to better performance. These findings provide preliminary evidence that a considerable amount of noise exists in the initial modality features and that not all encoded information contributes positively to the task. Consequently, applying an effective denoising strategy improves the quality of modality representations and ultimately enhances the overall performance of MSA task.

\begin{table}[t]

\includegraphics[width=\columnwidth]{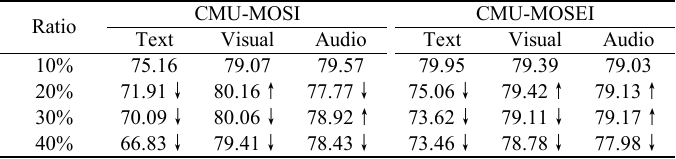}
\caption{Unimodal Noise. Text, visual, and audio represent the modality being masked, while the other modality features remain unchanged.}
\label{tab:unimodal_noise}
\end{table}

\subsection{Unimodal Noise}

To further investigate the impact of noise within different modalities, we apply masking to each individual modality separately. Using the MulT model as the base framework, we impose varying levels of masking on the text, audio, and visual modalities. We observe how model performance changes as the degree of masking increases for each single modality. This allows us to analyze the relative level of noise in the initial features of each modality and its effect on sentiment analysis.

As shown in Table~\ref{tab:unimodal_noise}, the audio and visual modalities exhibit performance trends that differ from those of the text modality. As the masking ratio increases, the performance of the audio and visual modalities shows fluctuating improvements. However, performance consistently declines in the text modality. This observation suggests that the initial text modality features are higher quality, containing less noise. Therefore, the text modality can serve as a reference standard. Moreover, the performance improvements in the audio and visual modalities are triggered at different masking ratios, further validating the different distribution of noise across modalities. This observation suggests that the differences in noise characteristics between different modalities may lead to the loss of critical information in some modalities when a unified denoising strategy is applied to all modalities. Therefore, differentiated denoising strategies should be adopted for different modalities.

\section{More Details}
\label{app:details}

\subsection{Dataset and Metrics}

These datasets encompass both Chinese and English corpora, incorporating text, audio, and visual modalities across diverse contexts such as monologues, dialogues, and film clips, with annotations covering both discrete and continuous emotional dimensions. For the CMU-MOSI and CMU-MOSEI datasets, we follow prior works \cite{wu2024multimodal,jiang2024joint,10.1145/3664647.3681527} to evaluate both regression and classification tasks. For regression, we report the \textbf{Mean Absolute Error (MAE)} and \textbf{Correlation coefficient (Corr)}. For classification, we calculate the \textbf{Acc$_2$} and \textbf{F1} scores for both the including zero sentiment scores as positive (ACC$_{2Has0}$ / F1$_{Has0}$) and the ignoring zero sentiment scores (ACC$_{2Non0}$ / F1$_{Non0}$). Additionally, we report \textbf{ACC$_7$}. For the CH-SIMS dataset, we adopt \textbf{ACC$_2$} and \textbf{F1} scores as evaluation metrics for the classification task. For the IEMOCAP dataset, we use \textbf{Weighted-F1} and \textbf{Macro-F1} scores to assess the performance of the classification task.

\subsection{Baselines}

\textbf{Framework:} We select influential and reproducible multimodal sentiment analysis models as comparative baselines:
\textbf{MulT} \cite{tsai2019multimodal}, \textbf{SPECTRA} \cite{yu-etal-2023-speech}, \textbf{KuDA} \cite{feng2024knowledge}, \textbf{SFTTR} \cite{sun-tian-2025-sequential}, \textbf{MMML} \cite{wu2024multimodal}. These models cover different research paradigms such as multimodal feature alignment, cross-modality attention modeling, and knowledge enhancement. With the rapid rise of the Multimodal Large Language Model (MLLM), unified understanding and reasoning across modalities has become a key direction in multimodal research. To verify the applicability and stability of our framework within this new paradigm, we further introduce \textbf{LLaVA-NeXT}\cite{li2024llava} and \textbf{Qwen2.5-VL}\cite{bai2025qwen2} as representative models for comparison, further comprehensively examining the framework's universality performance.

\noindent\textbf{Models:} In addition to the baseline models used in the framework experiments, we also compare a series of representative approaches for MSA:
\textbf{TFN} \cite{zadeh-etal-2017-tensor}: Models intra- and inter-modality dynamics in an end-to-end manner.
\textbf{LMF} \cite{liu-etal-2018-efficient-low}: Employs low-rank tensors for efficient multimodal fusion, reducing computational complexity.
\textbf{MFM} \cite{tsai2018learning}: Jointly optimizes generative and discriminative objectives on multimodal data and labels.
\textbf{Self-MM} \cite{Yu_Xu_Yuan_Wu_2021}: Designs a self-supervised label generation module to automatically obtain unimodal supervision signals.
\textbf{UniMSE} \cite{hu-etal-2022-unimse}: Unifies multimodal sentiment analysis and emotion recognition tasks from multiple perspectives.
\textbf{CHFN} \cite{10.1145/3503161.3548137}: Based on a Transformer architecture, it efficiently fuses unaligned multimodal sequences.
\textbf{ALMT} \cite{zhang-etal-2023-learning-language}: Introduces an adaptive hyper-modality learning module to suppress irrelevant or conflicting information under the guidance of language.
\textbf{EMT} \cite{sun2023efficient}: Enhances model robustness in scenarios with incomplete modalities while maintaining efficiency and performance.
\textbf{GLoMo} \cite{10.1145/3664647.3681527}: Integrates local representations from each modality and combines them with global representations to enhance expressive power.
\textbf{JOSFD} \cite{jiang2024joint}: Incorporates fuzzy logic to model both subjective and objective fuzziness in sentiment information.
\textbf{t-HNE} \cite{li2025t}: Text-guided hierarchical denoiser improves sentiment analysis performance via two-stage denoising and contrastive learning mechanism.
\textbf{t-HNE} is the \textbf{latest} MSA denoising method. \textbf{MMML} is the current \textbf{SOTA} MSA model.

\subsection{Implementation Detail}

All experiments are conducted using the PyTorch framework on a hardware setup with 8 RTX A6000 GPUs. For the framework experiments, we strictly follow the hyperparameter settings reported in the original papers to ensure fair and consistent comparisons. In our methodological experiments, to ensure fair comparison, we maintain the same parameter size settings for our modality encoders as in previous studies\cite{wu2024multimodal}. Model is trained using the AdamW optimizer to achieve stable and efficient convergence performance. Specifically, the learning rate is set to 5e-6 for the MOSI and MOSEI datasets, 1e-5 for the CH-SIMS dataset, and 2e-8 for the IEMOCAP dataset. In the framework experiments, we first fine-tune the MLLM using LoRA\cite{hu2022lora} on MSA datasets to adapt it to the task characteristics and stabilize model performance. We then integrate the fine-tuned model into the proposed framework to verify its effectiveness. In the method experiments, the MLLM results report the performance of the model after LoRA fine-tuning. To ensure the stability of the experimental results, we fix the random seed to 1 and run each experiment five times independently, reporting the average result.

\section{Significance Testing}
\label{app:st}
For Table~\ref{tab:framework_2}, we perform significance testing on seven experimental groups, with p-value of \( 4.26 \times 10^{-2} \), \( 6.01 \times 10^{-4} \), \( 2.95 \times 10^{-3} \), \( 1.08 \times 10^{-3} \), \( 1.19 \times 10^{-2} \), \( 4.28 \times 10^{-3} \), and \( 4.45 \times 10^{-3} \), all of which \(<\) 0.05 indicate significant differences.

For Table~\ref{tab:method_2}, the significance testing on MMML and MoLAN\textsuperscript{+}, with p-value (0.01 \(<\) 0.05)  indicating significant differences.

\section{More Experiments}

\label{app:more}

\begin{table}[t]

\includegraphics[width=\columnwidth]{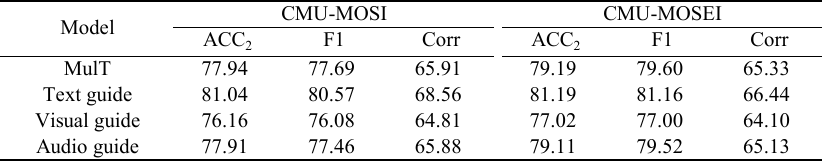}
\caption{Ablation study of denoising strength computation guidance.}
\label{tab:guide}
\end{table}

\subsection{Block Partitioning Strategy Selection} 

We adopt a block partitioning strategy based on the factor closest to the square root of the feature dimension, since not all feature dimensions are perfect squares. This design ensures a more balanced division of feature regions and avoids the two extremes of excessively large or excessively small blocks. Specifically, when the block is too large, it tends to mix effective information with noise within the same region, making it difficult for the denoising strength to accurately distinguish between the two. This may result in incomplete noise removal or excessive suppression of critical information. In contrast, when the block is too small, it over-segments semantic structures, weakens inter-block correlations. By comparison, the close to square root block partitioning strategy achieves a better trade-off between signal–noise separation and semantic integrity,  allowing the denoising process to retain critical information while removing noise.

As shown in Table \ref{tab:block}, we design comparative experiments with different block sizes for both visual and audio modalities. The experimental results show that, in both visual and audio feature processing, excessively large or small block partitioning lead to a decline in model performance. In contrast, using the close to square root block partitioning strategy achieves the best balance between critical information preservation and noise suppression, allowing the model to capture critical features more effectively and maintain overall performance stability. Overall, the proposed close to square root block partitioning strategy aligns well with the characteristics of each modality and lays a solid foundation for subsequent denoising.

\begin{figure}[t]
  \centering
  \includegraphics[width=\linewidth]{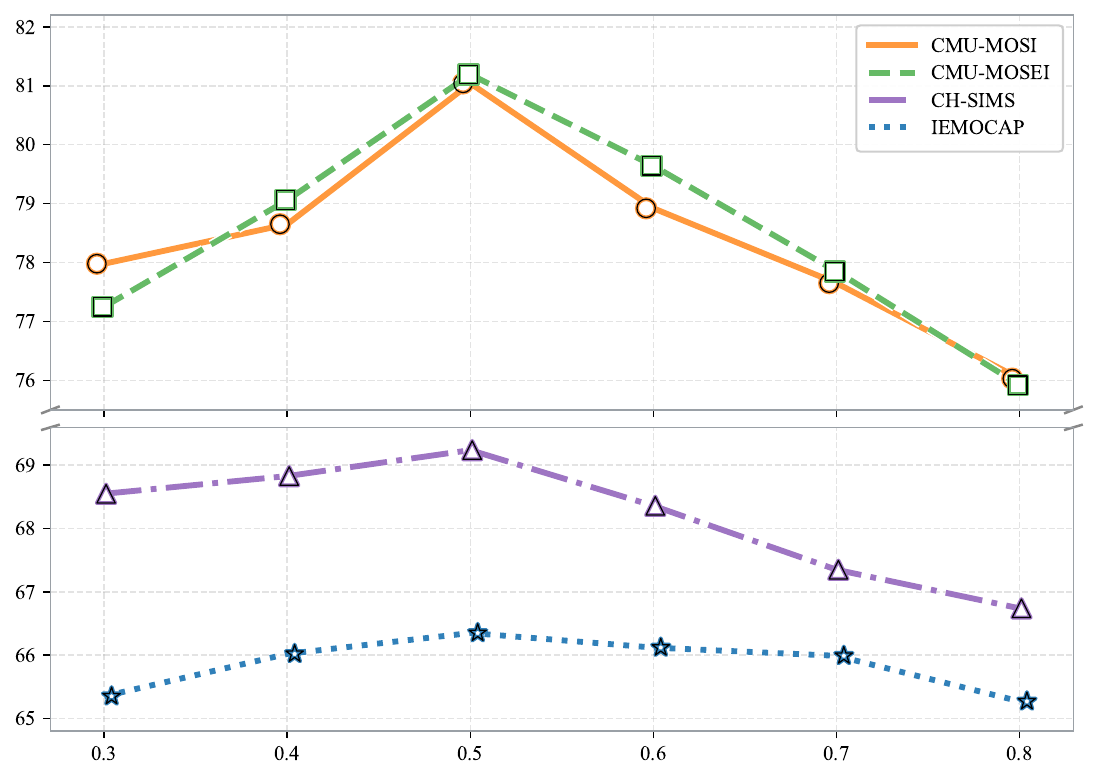}
  \caption{Performance comparison under different similarity thresholds $\theta$. The four curves represent experimental results on CMU-MOSI, CMU-MOSEI, CH-SIMS, and IEMOCAP datasets, respectively. The x-axis denotes the similarity threshold $\theta$, and the y-axis indicates the model performance.}
  \label{fig:diff_value}
\end{figure}

\subsection{Similarity Threshold Analysis}

To determine the optimal similarity threshold $\theta$ in the noise-suppressed cross attention mechanism, we conduct a series of comparative experiments across four benchmark datasets, as shown in Figure~\ref{fig:diff_value}. The threshold $\theta \in [0, 1]$ controls the activation of the denoising mask by determining which sub-blocks are preserved or suppressed based on similarity scores.

The experimental results demonstrate a consistent trend across all datasets: as $\theta$ increases from 0.3 to 0.8, model performance initially improves and then declines. A smaller $\theta$ (e.g., 0.3–0.4) allows excessive noisy features to pass through, resulting in incomplete denoising and suboptimal performance. Conversely, an overly large $\theta$ (e.g., 0.7–0.8) overly suppresses critical information, causing semantic loss and degraded performance. The performance peaks at ($\theta$ = 0.5), where the model achieves the best trade-off between noise suppression and information preservation. Therefore, we set $\theta = 0.5$ as the default similarity threshold in all experiments.

\begin{table}[t]

\includegraphics[width=\columnwidth]{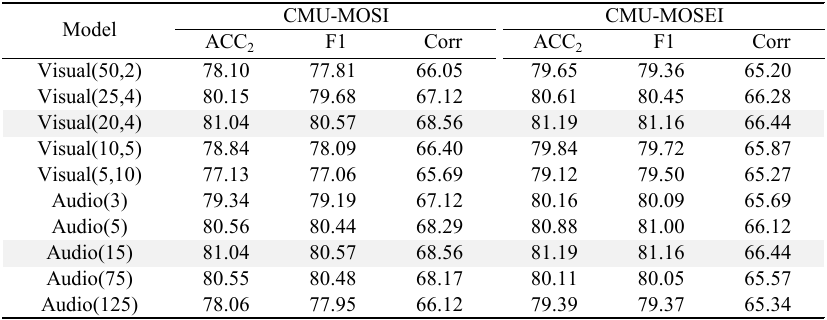}
\caption{Block Sizes Experiments.Take MulT+MoLAN on the CMU-MOSI and CMU-MOSEI dataset as an example.The audio feature dimension is $[128, 375, 20]$, and the visual feature dimension is $[128, 500, 20]$.The number after audio indicates the one-dimensional block size $(j)$, and the number after video indicates the two-dimensional block size $(k,j)$.}
\label{tab:block}
\end{table}

\subsection{Denoising-Driven Contrastive Learning Analysis}

To further verify the effectiveness of denoising-driven contrastive learning (DC) in modality denoising, we use two metrics in the field of contrastive learning: alignment and uniformity\cite{wang2020understanding}. Alignment measures the average distance between positive sample pairs in the embedding space, reflecting the model’s ability to cluster semantically similar samples. A smaller value indicates tighter intra-class compactness. Uniformity measures how evenly all samples are distributed on the unit hypersphere. A smaller value implies larger inter-class separation and better discrimination against noise.

As shown in Table ~\ref{tab:cl}, after incorporating DC, the model achieves lower alignment on both CMU-MOSI and CMU-MOSEI datasets, suggesting that positive samples cluster more tightly in the embedding space. Meanwhile, the uniformity score decreases significantly, indicating that negative samples become more separable and noise becomes easier to distinguish. These results indicate that DC effectively improves the model's ability to distinguish critical information features from noisy redundant features, enabling the MSA model to learn purer modality representations. 


\begin{table}[t]

\includegraphics[width=\columnwidth]{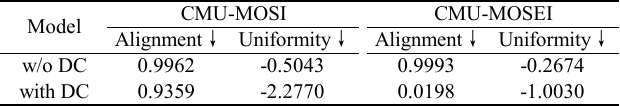}
\caption{Ablation studies of Denoising-Driven Contrastive Learning. Alignment and Uniformity metrics on CMU-MOSI and CMU-MOSEI datasets.}
\label{tab:cl}
\end{table}

\subsection{Denoising strength computation guidance} 
In our design, the denoising strength of each block is determined based on the text modality. To validate this choice, we conduct an ablation study comparing different guiding modalities, as shown in Table~\ref{tab:guide}. The results show that using text as the guidance yields the best performance on two datasets. In contrast, using visual or audio features as the guidance leads to a performance drop, indicating that these modalities introduce more noise. These findings are consistent with the results observed in pilot study (Table~\ref{tab:unimodal_noise}), further confirming that text-guided denoising offers more reliable representations for MSA.

\begin{figure}[!h]
  \centering
  \includegraphics[width=\linewidth]{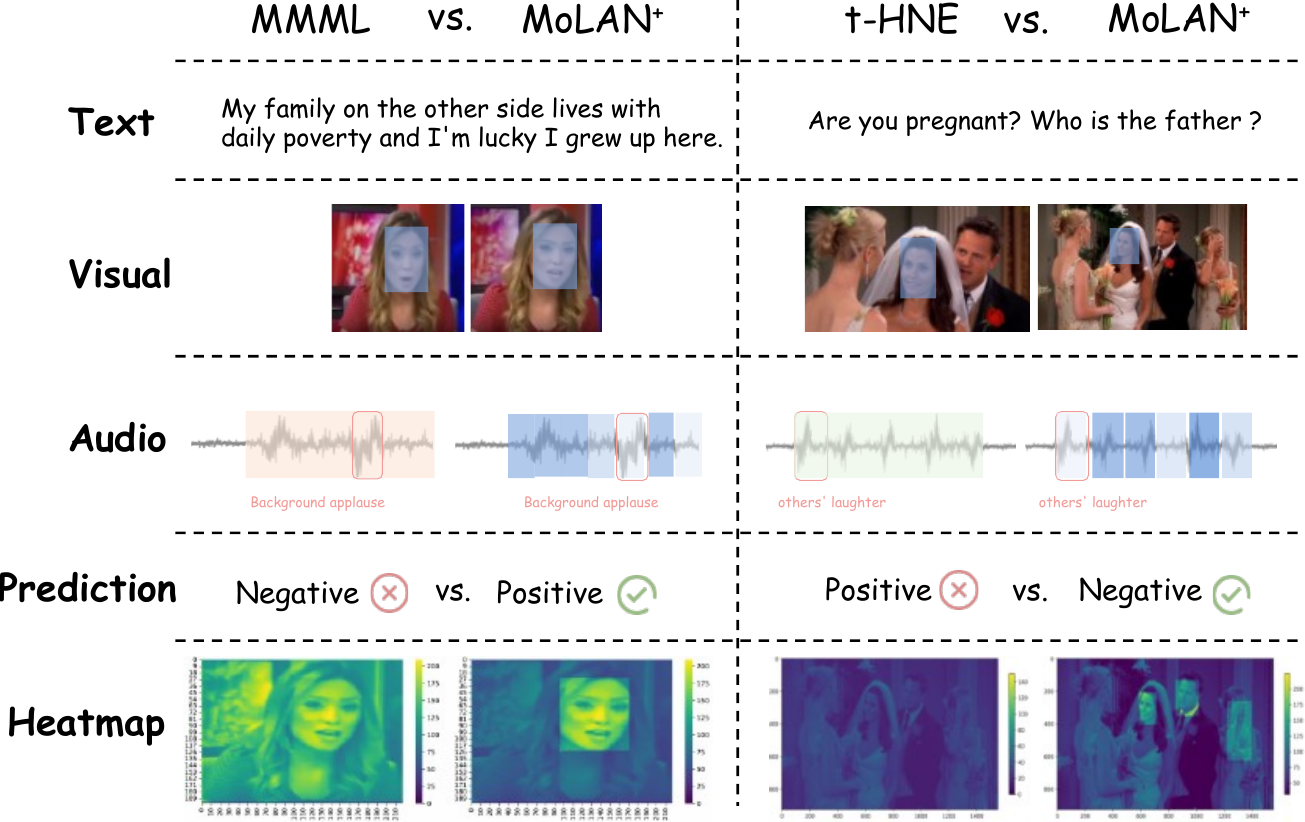}
  \caption{Case Study. The blue area in the visual modality represents the target person. The red, green, and blue colors in the audio modality represent the attention distribution of MMML, t-HNE, and MoLAN\textsuperscript{+} on different audio segments, respectively. The red boxes in the audio mark the noisy segments. The heatmap shows the model's attention strength in different visual regions.}
  \label{fig:case}
\end{figure}

\section{Case Study}
\label{app:case}
As shown in Figure~\ref{fig:case}, we present two representative case studies comparing MoLAN\textsuperscript{+} with the original SOTA model MMML and the latest denoising model t-HNE. In the left, we compare MoLAN\textsuperscript{+} with MMML. It can be observed that MMML pays almost equal attention to different visual and audio segments without effectively suppressing noise interference. Consequently, the model fails to distinguish the emotionally relevant regions and misinterprets the sentiment. In contrast, MoLAN\textsuperscript{+} focuses more accurately on the target speaker and the emotionally relevant regions, leading to a correct sentiment analysis. In the right, we compare MoLAN\textsuperscript{+} with t-HNE. Although t-HNE performs denoising, it applies a uniform denoising intensity across all modalities, which results in the loss of critical information necessary for MSA. This over-smoothing effect causes the model to generate incorrect sentiment. Conversely, MoLAN\textsuperscript{+} adaptively balances noise suppression and information retention through its dynamic noise-editing mechanism, thereby capturing emotion-relevant cues and achieving a more reliable analysis.

\end{document}